\documentclass[journal]{IEEEtran}
\usepackage{booktabs}
\usepackage{amsmath,amsfonts}
\usepackage{algorithmic}
\usepackage{array}
\usepackage{times}
\usepackage{helvet}  
\usepackage{courier}  
\usepackage[hyphens]{url}  
\usepackage{algorithm}
\usepackage{algorithmic}
\usepackage{newfloat}
\usepackage{listings}
\usepackage{multirow}
\usepackage[caption=false,font=normalsize,labelfont=sf,textfont=sf]{subfig}
\usepackage{textcomp}
\usepackage{stfloats}
\usepackage{url}
\usepackage{verbatim}
\floatstyle{ruled}
\newfloat{listing}{tb}{lst}{}
\floatname{listing}{Listing}

\usepackage{graphicx}
\def\BibTeX{{\rm B\kern-.05em{\sc i\kern-.025em b}\kern-.08em
    T\kern-.1667em\lower.7ex\hbox{E}\kern-.125emX}}
\usepackage{balance}
\usepackage{bm}  

\begin{document}
\title{Times2D: Multi-Period Decomposition and Derivative Mapping for General Time Series Forecasting}

\author{Reza Nematirad, Anil Pahwa, Balasubramaniam Natarajan
}

\maketitle

\begin{abstract}
Time series forecasting is an important application in various domains such as energy management, traffic planning, financial markets, meteorology, and medicine. However, real-time series data often present intricate temporal variability and sharp fluctuations, which pose significant challenges for time series forecasting. Previous models that rely on 1D time series representations usually struggle with complex temporal variations. To address the limitations of 1D time series, this study introduces the Times2D method that transforms the 1D time series into 2D space. Times2D consists of three main parts: first, a Periodic Decomposition Block (PDB) that captures temporal variations within a period and between the same periods by converting the time series into a 2D tensor in the frequency domain. Second, the First and Second Derivative Heatmaps (FSDH) capture sharp changes and turning points, respectively. Finally, an Aggregation Forecasting Block (AFB) integrates the output tensors from PDB and FSDH for accurate forecasting. This 2D transformation enables the utilization of 2D convolutional operations to effectively capture long and short characteristics of the time series. Comprehensive experimental results across large-scale data in the literature demonstrate that the proposed Times2D model achieves state-of-the-art performance in both short-term and long-term forecasting. The code is available in this repository: https://github.com/Tims2D/Times2D.
\end{abstract}

\section{Introduction}

Time series forecasting plays a critical role in several real-world applications, including electricity load demand forecasting, lifetime estimation of industrial machinery, traffic planning, weather prediction, and the stock market \cite{nematirad2025spdnetseasonalperiodicdecompositionnetwork}. Due to their critical relevance and wide application, there has been considerable interest in time series forecasting in recent years.  
The range of data resolutions in time series has historically been broad, from annual records such as GDP rates to monthly indicators like unemployment rates, and daily measurements such as stock market indices. However, advancements in data collection technologies have significantly increased the frequency of data recording. For example, observations range from hourly metrics like electricity use, to second-by-second updates in internet traffic and even millisecond-level data in rotary machines. Low-resolution time series data were typically smoother and exhibited fewer fluctuations, making prediction relatively straightforward. However, shifting toward high-resolution time steps introduces intricate temporal patterns and sharp fluctuations, making the time series forecasting task significantly challenging. 
Traditional methods such as Autoregressive Integrated Moving Average (ARIMA) and Exponential Smoothing were effective in handling well-defined seasonal patterns and trends. However, they struggle with intricate temporal and nonlinear patterns in time series data \cite{fu2022reinforcement}. Recently, due to advances in deep learning models, the landscape of time series forecasting has achieved promising progress. These models, by leveraging non-linear modeling, high-dimensional pattern recognition, and adaptive learning capabilities, have been able to capture complex temporal variations in real-world time series\cite{he2022catn}. Recurrent Neural Network (RNN) models are designed to process sequences by handling time points successively with internal state updates that capture temporal information. However, RNNs often struggle with long-term dependencies due to vanishing gradients \cite{massart2022coordinate}. Long Short-Term Memory (LSTM) models were proposed to address the vanishing gradient problem. However, LSTMs have difficulty capturing local trends \cite{huang2024mdg}. Temporal Convolutional Networks (TCNs) are utilized to extract the variation information by applying convolutional neural networks. TCNs are efficient at capturing relationships between adjacent time points but may struggle to model long-term dependencies \cite{wu2022timesnet}. Recently, transformer-based models have shown superior performance over TCN models due to their attention mechanism. However, the attention mechanism can struggle with capturing dependencies among scattered time points \cite{wu2021autoformer} and can be computationally extensive, especially for long-term forecasting \cite{dai2024periodicity}. 

It should be noted that intricate temporal variations in real-world data often manifest as sharp fluctuations, abrupt rises and falls, and overlapping patterns. Additionally, most real-world datasets exhibit multi-periodicity, which further complicates the modeling process. These periods may overlap, creating additional challenges when attempting to model them using a 1D perspective. Motivated by the limitations of the existing deep learning models and the complex characteristics observed in real-world time series data, this study introduces a novel framework for general time series forecasting. The proposed approach specifically addresses the challenges of multi-periodicity, sharp fluctuations, and turning points in time series data. Technically, Times2D consists of three major components. First, the Periodic Decomposition Block (PDB) identifies the top $k$ dominant periods and their corresponding frequencies using the Fast Fourier Transform (FFT). For each dominant period-frequency pair, the 1D time series is reshaped into a 2D tensor, where the rows represent the period and the columns correspond to the frequency. As a result, the 1D input time series is transformed into $k$ distinct 2D tensors. This transformation facilitates the model ability to capture both short-term and long-term dependencies. Second, the First and Second Derivative Heatmaps (FSDH) operate separately from the PDB. This component calculates the first and second derivatives of the original 1D time series, constructing 2D tensors that highlight sharp changes and turning points in the data. Finally, the Aggregation Forecasting Block (AFB) integrates the outputs generated by the PDB and FSDH to produce accurate forecasts, effectively capturing the intricate temporal variations inherent in the time series. Figure \ref{Figure1} provides an overview of the proposed Times2D framework.
The contributions of this study are summarized as follows:
\begin{figure*}[h]
\centering
\includegraphics[width=0.93\textwidth, keepaspectratio]{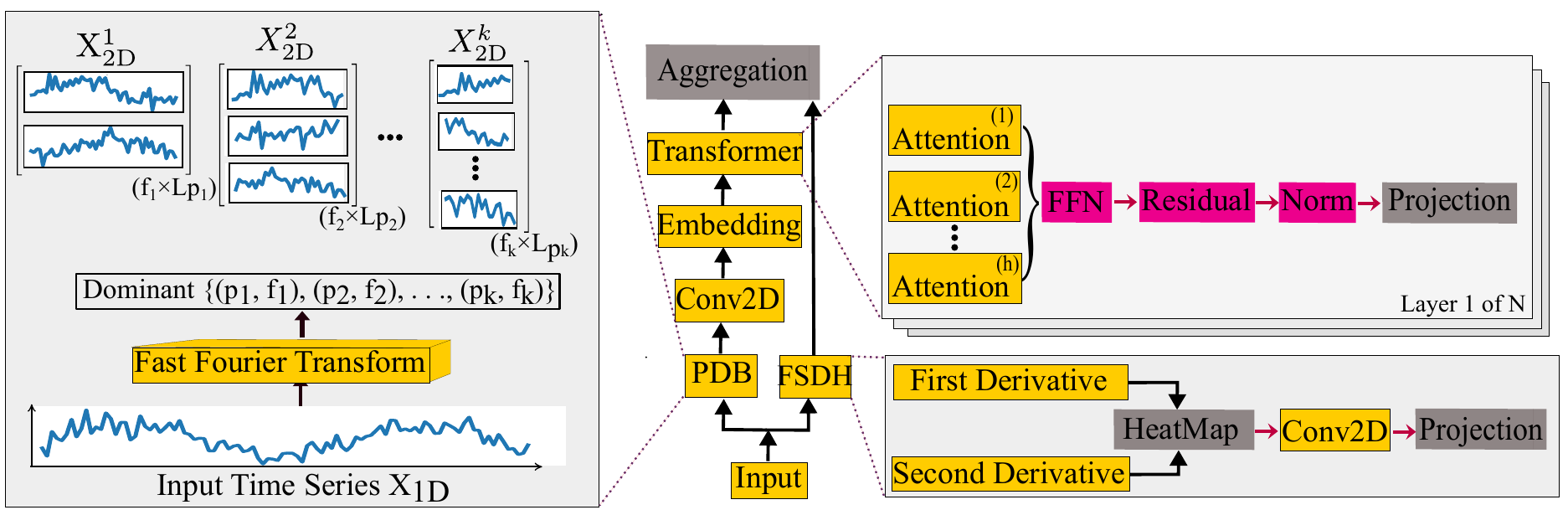} 
\caption{Overview of the proposed Times2D framework for general time series forecasting.}
\label{Figure1}
\end{figure*}

\begin{itemize}
\item This study introduces the Times2D framework, which transforms 1D time series data into 2D space to capture complex temporal patterns, including multi-periodicity, sharp fluctuations, and turning points for general time series forecasting.
\item We propose the PDB, which decomposes the time series into segments based on dominant periods identified through FFT. This block effectively captures both short-term and long-term dependencies.
\item The study introduces the FSDH block, which calculates the first and second derivatives to emphasize sharp changes and turning points in the time series.
\item Comprehensive experiments on various large-scale datasets demonstrate that the proposed Times2D model achieves state-of-the-art performance in both short-term and long-term forecasting.
\end{itemize}

\section{Related Work}
Classical methods such as ARIMA and Exponential Smoothing aim to extract predefined temporal variations, such as trends and seasonality. While these methods are effective for capturing well-defined seasonal patterns and trends, they often struggle with more complex, non-linear patterns that are present in real-world time series data \cite{zhang2024sageformer}. In recent years, deep learning has made significant contributions to time series forecasting. Based on the literature review in this field, deep learning models to capture temporal variations in time series data can be categorized into four groups.

\subsection{Recurrent Neural Networks}
RNNs and their variants, LSTMs and Gated Recurrent Units (GRUs), have been widely adopted for modeling sequential data. RNNs process sequences by maintaining a hidden state that captures temporal dependencies \cite{tan2023openstl}. However, they often suffer from vanishing gradients, which makes it difficult for RNNs to capture long-term relationships \cite{liu2022pyraformer}. LSTM networks overcome this issue by introducing memory cells and gate mechanisms to control the flow of information. This modification enables LSTMs to capture long-term dependencies. An example of an LSTM-based model is proposed in \cite{li2023extreme} to predict 3-day ahead hourly water level.  However, LSTMs still face challenges in capturing local trends and rapid fluctuations.
\subsection{Convolutional Neural Networks}
TCN models in time series analysis take advantage of causal convolutions, dilated convolutions, and residual connections. Causal convolutions ensure that outputs at time step t depend only on inputs from time step t and earlier. This aspect of TCNs prevents future data leakage, which is essential for accurate time series forecasting. Dilated convolutions expand the receptive field exponentially without increasing computational complexity. This enhances the ability of the model to capture long-term dependencies. Residual connections mitigate the vanishing gradient problem. They enable the training of deeper networks by providing direct paths for gradient flow \cite{wang2023micn}. A TCN is developed in \cite{ahmadi2014hybrid} to predict the weather, and experimental results show that the TCN outperforms the LSTM and other classical machine learning methods in predicting weather. A Multi-scale Iso-metric Convolution Network (MICN) is proposed in \cite{wang2023micn} to efficiently capture the local and global patterns in time series data for long-term forecasting. Temporal 2D-Variation Modeling to capture the intricate temporal variations in time series data analysis is proposed in \cite{wu2022timesnet}. Despite the advantages of TCNs, the locality of one-dimensional convolution kernels and receptive fields restricts their ability to capture relationships between distant time points \cite{luo2024moderntcn}.
\subsection{Transformers}
Transformers have revolutionized time series forecasting models with their powerful attention mechanisms. Unlike RNNs and TCNs, transformers use a self-attention mechanism to weigh the importance of time steps. Accordingly, they can effectively capture long-term dependencies and relationships between distant time points \cite{zeng2023transformers}. Pyramidal Attention Module (PAM) is introduced in \cite{liu2022pyraformer} to model temporal dependencies with low complexity in long-term forecasting tasks. In Autoformer model \cite{wu2021autoformer}, a novel decomposition architecture with an Auto-Correlation mechanism is designed to effectively capture the seasonal and trend patterns of time series data. Besides, to further enhance the seasonal and trend decomposition, a sparse attention within the frequency domain is designed as the Frequency Enhanced Decomposed Transformer (FEDformer) \cite{zhou2022fedformer}, which enhances the performance of Transformers for long-term forecasting. Despite their strengths, it is challenging for the attention mechanism to reliably discern dependencies directly from scattered time points because temporal dependencies can be deeply hidden by intricate temporal patterns \cite{wu2022timesnet}. 
\subsection{Multi-layer Perceptron}
Multi-layer Perceptrons (MLPs) are fundamental neural network architectures consisting of fully connected layers. A deep neural architecture called N-BEATS is proposed by \cite{oreshkin2019n} for univariate time series forecasting. The architecture uses backward and forwards residual links and a very deep stack of fully-connected layers. A lightweight deep learning architecture called LightTS is introduced by \cite{zhang2022less} for multivariate time series forecasting. LightTS utilizes simple MLP-based structures and incorporates two down-sampling strategies, interval sampling, and continuous sampling, to preserve the majority of the time series information. Besides, to address volatility and computational complexity in long-horizon forecasting, Neural Hierarchical Interpolation for Time Series Forecasting (NHITS) is introduced in \cite{challu2023nhits}. 

Although valuable studies have been done in time series forecasting tasks, we reveal intricate temporal information within time series data by employing multi-periodicity and derivative representations to explore temporal patterns in 2D space. In this transformation from 1D to 2D, 2D kernels are able to capture various complex temporal variations such as periodicity, rising, falling, sharp fluctuations, and turning points for the first time.

\section{Times2D}
 Given a multivariate time series dataset \( X = [x_1, x_2, \dots, x_T] \in \mathbb{R}^{T \times N} \), where \( T \) represents the length of the time series and \( N \) denotes the number of variables or dimensions. The dataset must be appropriately formatted by defining the input and output sequences. First, a segment of the time series, referred to as the sequence length \( S \), is selected as the input. A subsequent segment of the time series, referred to as the prediction length \( P \), is selected as the output. By following this approach, the original time series of total length \( T \) is transformed into multiple rows, where each row contains \( S \) input time points and \( P \) corresponding output. Additionally, the data is organized into batches for parallel processing during training. As a result, the samples are organized into batches of size \( B \), resulting in input tensors with dimensions \( [B, S, N] \) and output tensors with dimensions \( [B, P, N] \). 

\subsection{Periodic Decomposition Block}
The PDB identifies dominant periods within the time series data in frequency domain. 
Given the input tensor \( X_{\text{1D}} \)  with dimensions \( [B, S, N] \), the FFT can be calculated as:
\begin{equation}
X_f =  \sum_{t=0}^{S-1} X(t) e^{-2 \pi i t f / S}
\end{equation}
where \( X_f \) represents the transformed tensor in the frequency domain and \( f \) is the frequency. The magnitude \( A_f \) can be calculated as follows: 
\begin{equation}
A_f = |X_f| = \sqrt{\text{Re}(X_f)^2 + \text{Im}(X_f)^2}
\end{equation}
Since \( X_f \) is symmetric, we maintain only the first half of the frequency components, corresponding to the non-negative frequencies. Accordingly, the magnitude \( A_f \) becomes a tensor of dimensions \( B \times \frac{S}{2} \times N \). To identify the dominant periods for each sequence, the magnitudes are averaged over the \( B \) and \( N \) dimensions. The resulting magnitudes are then sorted in descending order, and the dominant frequencies are selected based on their associated magnitudes. As a result, the dominant periods \( p_1, p_2, \dots, p_k \), associated with the dominant frequencies \( f_1, f_2, \dots, f_k \), can be calculated as:
\begin{equation}
P_k = \frac{S}{f_k}
\end{equation}
Thus, the original 1D input tensor \( X \in \mathbb{R}^{B \times S \times N} \) can be decomposed into \( k \) distinct 2D tensors \( X_{\text{2D}}^i \in \mathbb{R}^{B \times p_i \times f_i \times N} \), each corresponding to one of the identified dominant periods. Then the tensors  \( X_{\text{2D}}^i\) are passed through 2D convolutional layers to extract intricate relationships within and between periods. This convolution operates across both the periodic dimension \( p_i \) and the frequency dimension \( f_i \) as follows:
\begin{equation}
\hat{X}_{\text{2D}}^i = \text{Conv2D}(X_{\text{2D}}^i)
\end{equation}
Where \(\hat{X}_{\text{2D}}^i\) represents the output features after applying the convolutional kernel. Subsequently, the 2D tensors \(\hat{X}_{\text{2D}}^i\) are fed into the following mechanisms:

\subsubsection{Positional Embedding (PosEnc):} PosEnc encodes positional information to maintain the temporal and spatial structure within the 2D tensors.
\begin{equation}
{\hat{\bm{X}}}_{2D}^i=PosEnc({\hat{\bm{X}}}_{2D}^i)
\end{equation}
\subsubsection{Multi-Head Self-Attention (MHSA):} MHSA Captures intricate dependencies within the 2D tensors by considering relationships across spatial and temporal dimensions.
\begin{equation}
{\hat{\bm{X}}}_{2D}^i=MHSA({\hat{\bm{X}}}_{2D}^i)
\end{equation}
\subsubsection{Feed-Forward Networks (FFN):} FFN applies non-linear transformations to enhance the ability of the model to learn complex patterns in the data.
\begin{equation}
{\hat{\bm{X}}}_{2D}^i=FFN({\hat{\bm{X}}}_{2D}^i)
\end{equation}

\subsubsection{Residual Connection (Res):} A residual connection is applied to preserve the original information from the 2D tensors and facilitate gradient flow during training. \begin{equation} {\hat{\bm{X}}}_{2D}^i = {\hat{\bm{X}}}_{2D}^i + Res({\hat{\bm{X}}}_{2D}^i) \end{equation}

\subsubsection{Normalization (Norm):} Norm normalizes the tensor to stabilize training and improve convergence. \begin{equation} {\hat{\bm{X}}}_{2D}^i = Norm({\hat{\bm{X}}}_{2D}^i) \end{equation}
The normalized outputs are then transformed back into a 1D representation and the desired prediction length by applying a linear transformation as: 
\begin{equation}
{\hat{\bm{X}}}_{1D}^{PDB}=Linear({\hat{\bm{X}}}_{2D}^i)
\end{equation}
Where \({\hat{\bm{X}}}_{1D}^{PDB}\), with shape \([B, P, N]\), denotes the processed 1D representation obtained from the PDB, which serves as the input to subsequent layers for generating the final forecasted values.

\subsection{First and Second Derivative Heatmaps}

\begin{figure}[t]
\centering
\includegraphics[width=0.95\columnwidth]{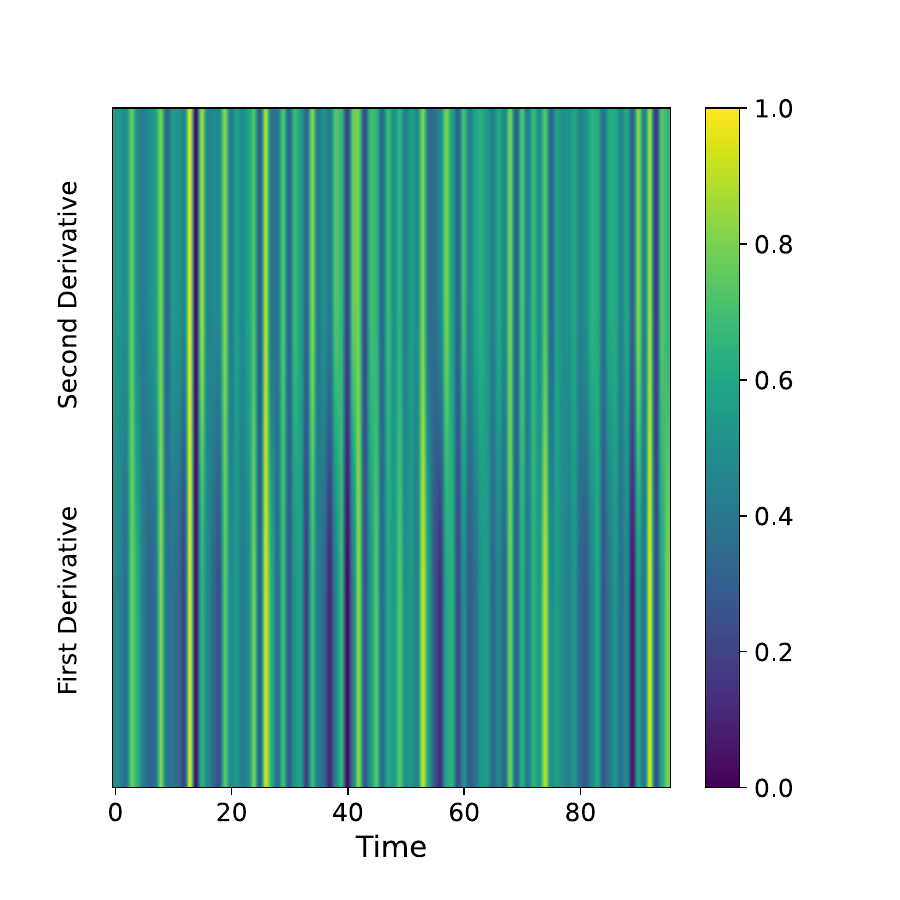} 
\caption{2D heatmap representation of a 1D sequence with a length of 96.}
\label{fig2}
\end{figure}

To effectively capture sharp fluctuations and turning points within the time series data, we compute the first and second derivatives. Given the input tensor \( X_{\text{1D}} \) with dimensions \([B, S, N ]\), the first derivative tensor \( D_{\text{1}} \) can be calculated as:
\begin{equation}
\bm{D}_1\left(t\right)=\ \bm{X}_{1D}\left(t+1\right)-\bm{X}_{1D}\left(t\right)
\end{equation}
To maintain the original sequence length, the first derivative tensor is padded at the beginning with zeros. Next, the second derivative tensor \( D_{\text{2}} \) is computed from the first derivative tensor as:
\begin{equation}
\bm{D}_2\left(t\right)=\ \bm{D}_1\left(t+1\right)-\bm{D}_1\left(t\right)
\end{equation}
Similarly, the second derivative tensor is padded with zeros at the beginning. To construct the final 2D heatmap tensor \( H_{\text{2D}}\) the padded first and second derivative tensors are stacked along a new dimension as:
\begin{equation}
\bm{H}_{2D}(t,D)=\ Stack\left(\bm{D}_1\left(t\right),\ \bm{D}_2\left(t\right)\right)
\end{equation}
For example, Figure~\ref{fig2} demonstrates a 2D heatmap representation of a 1D sequence with a length of 96. These derivatives are 2D representations where the X-axis represents time, and the Y-axis is conceptually divided into two parts, with the lower half representing the first derivative (indicating sharp fluctuations) and the upper half representing the second derivative (indicating turning points). The colors in the heatmap correspond to the magnitude of these derivatives, with more intense colors indicating greater changes. Then, to enhance the ability of the model to extract meaningful temporal features, the 2D derivative heatmap tensor is processed through a series of convolutional layers.

\begin{equation}
{\hat{\bm{H}}}_{2D}\left(t,D\right)=\ Conv2D(\bm{H}_{2D}(t,D))
\end{equation}
Finally, the extracted features are transformed back into a 1D representation and the desired prediction length by applying a learned set of weights and performing a weighted summation across the derivative dimension as:
\begin{equation}
{\hat{\bm{X}}}_{1D}^{FSDH}=\ Sum({\hat{\bm{H}}}_{2D}\left(t,D\right).weigths)
\end{equation}
Where \(\hat{\bm{X}}_{1D}^{FSDH}\), with shape \([B, P, N]\), denotes the processed 1D representation obtained from the FSDH block, which serves as the input to subsequent layers for generating the final forecasted values.
\begin{table*}[t]
\centering
\caption{Short-term forecasting results on the M4 dataset with prediction lengths of 6 for yearly, 8 for quarterly, and 18 for monthly. \textbf{Bold} and \underline{underlined} values indicate the best and second-best performance, respectively.}
\fontsize{8.7}{10}\selectfont 
\setlength{\tabcolsep}{0.3mm} 
\renewcommand{\arraystretch}{1.1} 
\begin{tabular}{c|c|c|c|c|c|c|c|c|c|c|c|c|c|c c}
\toprule
\multirow{2}{*}{Data} & \multirow{2}{*}{Metric} 
& Times2D & N-HITS & PatchTST & DLinear & TimesNet & MICN & SCINet & FEDformer & Stationary & FiLM & LightTS & Autoformer & Informer \\ 
& & (Ours) & (2023) & (2023) & (2023) & (2023) & (2023) & (2023) & (2023) & (2022) & (2022) & (2022) & (2021) & (2021) \\ 
\midrule
\multirow{3}{*}{\rotatebox{90}{Yearly}} 
& SMAPE & \textbf{13.418} & \underline{13.422} & 13.477 & 15.11 & 15.378 & 25.022 & 18.605 & 13.728 & 13.717 & 17.431 & 14.247 & 13.974 & 14.727 \\
& MASE & \textbf{2.990} & 3.056 & \underline{3.019} & 3.565 & 3.554 & 7.162 & 4.471 & 3.048 & 3.078 & 4.043 & 3.109 & 3.134 & 3.418 \\
& OWA & \textbf{0.787} & 0.795 & \underline{0.792} & 0.911 & 0.918 & 1.667 & 1.132 & 0.803 & 0.807 & 1.042 & 0.827 & 0.822 & 0.881 \\
\midrule
\multirow{3}{*}{\rotatebox{90}{Quarterly}}
& SMAPE & \underline{10.362} & \textbf{10.185} & 10.380 & 10.597 & 10.465 & 15.214 & 14.871 & 10.792 & 10.958 & 12.925 & 11.364 & 11.338 & 11.360 \\
& MASE & \underline{1.221} & \textbf{1.180} & 1.233 & 1.253 & 1.227 & 1.963 & 2.054 & 1.283 & 1.325 & 4.664 & 1.328 & 1.365 & 1.401 \\
& OWA & \underline{0.915} & \textbf{0.893} & 0.921 & 0.938 & 0.923 & 1.407 & 1.424 & 0.958 & 0.981 & 1.193 & 1.000 & 1.012 & 1.024 \\
\midrule
\multirow{3}{*}{\rotatebox{90}{Monthly}}
& SMAPE & \textbf{12.918} & 13.059 & \underline{12.959} & 13.258 & 13.513 & 16.943 & 14.925 & 14.260 & 13.917 & 15.407 & 14.014 & 13.958 & 14.062 \\
& MASE & \textbf{0.959} & 1.013 & \underline{0.970} & 1.003 & 1.039 & 1.442 & 1.131 & 1.102 & 1.097 & 1.298 & 1.053 & 1.103 & 1.141 \\
& OWA & \textbf{0.898} & 0.929 & \underline{0.905} & 0.931 & 0.957 & 1.265 & 1.027 & 1.012 & 0.998 & 1.144 & 0.981 & 1.002 & 1.024 \\
\bottomrule
\end{tabular}

\label{Table1}
\end{table*}

\subsection{Aggregation Block}
After obtaining the 1D representations \(\hat{\bm{X}}_{1D}^{PDB}\) from the PDB and \(\hat{\bm{X}}_{1D}^{FSDH}\) from the FSDH, these two components are aggregated to create a combined feature representation for the final forecasting task. This is achieved by performing an element-wise summation of the two 1D representations:
\begin{equation}
{\hat{\bm{X}}}_{1D}=\ {\hat{\bm{X}}}_{1D}^{PDB}+\ {\hat{\bm{X}}}_{1D}^{FSDH}
\end{equation}
The aggregated representation, \({\hat{\bm{X}}}_{1D} \in \mathbb{R}^{B \times P \times N}\), encapsulates various temporal variations captured by the PDB and FSDH. This combined representation is subsequently passed through a loss function during the training phase to optimize the model parameters for accurate forecasting.

\section{Experimental Results}
\subsection{Datasets}
We evaluated the Times2D model using several publicly available datasets commonly used in time series forecast-ing. For long-term forecasting, we utilized datasets like the electricity transformer temperature (ETT) and transporta-tion traffic data. These datasets cover periods ranging from July 2016 to July 2018 with frequencies of 15 minutes (ETTm1 and ETTm2) to 1 hour (ETTh1, ETTh2, and Traffic). For short-term forecasting, we employed sub-sets of the M4 dataset, which include time series from finance, economics, demographics, and industry, with frequencies ranging from monthly to yearly \cite{jin2023time}.

\subsection{Evaluation Metrics}
To comprehensively evaluate the performance of the Times2D, we used mean square error (MSE) and mean ab-solute error (MAE) for long-term forecasting, and symmetric mean absolute percentage error (SMAPE), mean absolute scaled error (MASE), and overall weighted average (OWA) for short-term forecasting. These metrics can be calculated as follows:
\begin{equation}
\mathrm{MSE} = \frac{1}{H} \sum_{i=1}^{H} (\hat{\mathbf{Y}}_i - \mathbf{Y}_i)^2
\end{equation}

\begin{equation}
\mathrm{MAE} = \frac{1}{H} \sum_{i=1}^{H} \left| \hat{\mathbf{Y}}_i - \mathbf{Y}_i \right|
\end{equation}

\begin{equation}
\mathrm{SMAPE} = \frac{200}{H} \sum_{i=1}^{H} \frac{\left| \hat{\mathbf{Y}}_i - \mathbf{Y}_i \right|}{\left| \hat{\mathbf{Y}}_i \right| + \left| \mathbf{Y}_i \right|}
\end{equation}

\begin{equation}
\mathrm{MASE} = \frac{\frac{1}{H} \sum_{i=1}^{H} \left| \hat{\mathbf{Y}}_i - \mathbf{Y}_i \right|}{\frac{1}{H-s} \sum_{j=s+1}^{H} \left| \mathbf{Y}_j - \mathbf{Y}_{j-s} \right|}
\end{equation}

\begin{equation}
\mathrm{OWA} = \frac{1}{2} \left( \frac{\mathrm{SMAPE}}{\mathrm{SMAPE}_{\text{naive}}} + \frac{\mathrm{MASE}}{\mathrm{MASE}_{\text{naive}}} \right)
\end{equation}
Where $\hat{\mathbf{Y}}_i$ and $\mathbf{Y}_i$ denote the predicted and actual values for the $i$-th observation, $H$ represents the forecasting horizon, $s$ is the seasonal period length, and $\mathrm{SMAPE}_{\text{naive}}$ and $\mathrm{MASE}_{\text{naive}}$ refer to the SMAPE and MASE values obtained using a naive forecasting method \cite{jin2023time}.

\subsection{Baselines}
The performance of Times2D is compared with multiple state-of-the-art algorithms in the time series forecasting domain. These models are categorized into different model architectures. Transformer-based models include PatchTST (2023), FEDformer (2022b), Stationary (2022b), Autoformer (2021), Informer (2021), CARD (2024), Sage-Former (2024), and Crossformer (2023). MLP-based models encompass N-HITS (2023), DLinear (2023), and LightTS (2022). TCN-based models include TimesNet (2023), MICN (2023), and SCINet (2022a). Additionally, the comparison includes GPT4TS from the category of large language models (LLMs). 

\begin{table*}[t]
\centering
\caption{Long-term forecasting results with prediction horizons of $H = \{96, 192, 336, 720\}$ for all experiments. \textbf{Bold} and \underline{underlined} values indicate the best and second-best performance, respectively.}
\fontsize{8.7}{10}\selectfont 

\setlength{\tabcolsep}{0.9mm} 

\begin{tabular}{c|c|cc|cc|cc|cc|cc|cc|cc|cc|cc}
\toprule
\multirow{3}{*}{\rotatebox{90}{Models}} & \multirow{3}{*}{$H$} & \multicolumn{2}{c|}{Time2D} & \multicolumn{2}{c|}{CARD} & \multicolumn{2}{c|}{SageFormer} & \multicolumn{2}{c|}{N-HITS} & \multicolumn{2}{c|}{GPT4TS} & \multicolumn{2}{c|}{PatchTST} & \multicolumn{2}{c|}{DLinear} & \multicolumn{2}{c|}{TimesNet} & \multicolumn{2}{c}{Crossformer} \\
 &  & \multicolumn{2}{c|}{(Ours)} & \multicolumn{2}{c|}{(2024)} & \multicolumn{2}{c|}{(2024)} & \multicolumn{2}{c|}{(2023)} & \multicolumn{2}{c|}{(2023)} & \multicolumn{2}{c|}{(2023)} & \multicolumn{2}{c|}{(2023)} & \multicolumn{2}{c|}{(2023)} & \multicolumn{2}{c}{(2023)} \\
 &  & MSE & MAE & MSE & MAE & MSE & MAE & MSE & MAE & MSE & MAE & MSE & MAE & MSE & MAE & MSE & MAE & MSE & MAE \\

\midrule
\multirow{5}{*}{\rotatebox{90}{ETTh1}} & 96  & \textbf{0.362} & \textbf{0.395} & 0.383 & 0.391 & 0.377 & 0.397 & 0.378 & 0.393 & 0.376 & \underline {0.397 }& \underline{0.370} & 0.399 & 0.375 & 0.399 & 0.384 & 0.402 & 0.423 & 0.448 \\
 & 192 & \textbf{0.405} & 0.423 & 0.435 & 0.420 & 0.423 & 0.425 & 0.427 & 0.436 & 0.416 & \underline{0.418} & 0.413 & \textbf{0.416} & \textbf{0.405} & \textbf{0.416} & 0.436 & 0.429 & 0.471 & 0.474 \\
 & 336 & \underline{0.423} & \textbf{0.433} & 0.479 & 0.442 & 0.459 & 0.445 & 0.458 & 0.484 & 0.442 & \textbf{0.433} & \textbf{0.422} & 0.436 & 0.439 & 0.443 & 0.491 & 0.469 & 0.570 & 0.546 \\
 & 720 & \textbf{0.445} & 0.465 & 0.471 & \textbf{0.461} & 0.465 & 0.466 & 0.472 & 0.561 & 0.477 & 0.456 & \underline{0.447} & \underline{0.466} & 0.472 & 0.490 & 0.521 & 0.500 & 0.653 & 0.621 \\
 & Avg & \textbf{0.410} & \textbf{0.429} & 0.442 & \textbf{0.429} & 0.431 & 0.433 & 0.433 & 0.468 & 0.465 & 0.455 & \underline{0.413} & \textbf{0.429} & 0.422 & 0.437 & 0.458 & 0.450 & 0.529 & 0.522 \\
\midrule
\multirow{5}{*}{\rotatebox{90}{ETTh2}} & 96  & \textbf{0.270} & \underline{0.334} & 0.281 & \textbf{0.330} & 0.286 & 0.338 & \underline{0.274} & 0.345 & 0.285 & 0.342 & \underline{0.274} & 0.336 & 0.289 & 0.353 & 0.340 & 0.374 & 0.745 & 0.584 \\
 & 192 & \textbf{0.334} & \textbf{0.378} & 0.363 & 0.381 & 0.368 & 0.394 & 0.353 & 0.401 & 0.354 & 0.389 & \underline{0.339} & \underline{0.379} & 0.383 & 0.418 & 0.402 & 0.414 & 0.877 & 0.656 \\
 & 336 & \textbf{0.329} & \underline{0.385} & 0.411 & 0.418 & 0.413 & 0.429 & 0.382 & 0.425 & \underline{0.373} & 0.407 & \textbf{0.329} & \textbf{0.380} & 0.448 & 0.465 & 0.452 & 0.452 & 1.043 & 0.731 \\
 & 720 & \textbf{0.376} & \textbf{0.422} & 0.416 & 0.431 & 0.427 & 0.449 & 0.625 & 0.557 & 0.406 & 0.441 & \underline{0.379} & \underline{0.422} & 0.605 & 0.551 & 0.462 & 0.468 & 1.104 & 0.763 \\
 & Avg & \textbf{0.327} & \textbf{0.379} & 0.368 & \underline{0.390} & 0.374 & 0.403 & 0.408 & 0.507 & 0.381 & 0.412 & \underline{0.330} & \textbf{0.379} & 0.431 & 0.446 & 0.414 & 0.427 & 0.942 & 0.684 \\
\midrule
\multirow{5}{*}{\rotatebox{90}{ETTm1}} & 96  & \textbf{0.279} & \textbf{0.343} & 0.316 & 0.347 & 0.324 & 0.362 & 0.302 & 0.350 & 0.292 & 0.346 & \underline{0.290} & \underline{0.342} & 0.299 & 0.343 & 0.338 & 0.375 & 0.404 & 0.426 \\
 & 192 & \textbf{0.313} & \textbf{0.361} & 0.363 & 0.370 & 0.368 & 0.387 & 0.374 & 0.383 & 0.332 & 0.372 & \underline{0.332} & 0.369 & 0.335 & \underline{0.365} & 0.327 & 0.387 & 0.450 & 0.451 \\
 & 336 & \textbf{0.347} & \textbf{0.380} & 0.392 & 0.390 & 0.401 & 0.408 & 0.369 & 0.402 & \underline{0.366} & 0.394 & \underline{0.366} & 0.392 & 0.369 & \underline{0.386} & 0.410 & 0.411 & 0.532 & 0.515 \\
 & 720 & \textbf{0.410} & \textbf{0.413} & 0.458 & 0.425 & 0.457 & 0.441 & 0.431 & 0.441 & 0.417 & 0.421 & \underline{0.416} & \underline{0.420} & 0.425 & 0.421 & 0.478 & 0.450 & 0.666 & 0.589 \\
 & Avg & \textbf{0.337} & \textbf{0.374} & 0.383 & 0.384 & 0.388 & 0.400 & 0.369 & 0.394 & 0.388 & 0.403 & \underline{0.351} & 0.380 & 0.357 & \underline{0.378} & 0.400 & 0.406 & 0.513 & 0.495 \\
\midrule
\multirow{5}{*}{\rotatebox{90}{ETTm2}} & 96  & \textbf{0.166} & 0.259 & 0.169 & \textbf{0.248} & 0.173 & 0.255 & 0.176 & 0.255 & 0.173 & 0.262 & \textbf{0.166} & \underline{0.255} & \underline{0.167} & 0.269 & 0.187 & 0.267 & 0.287 & 0.366 \\
 & 192 & \textbf{0.220} & \underline{0.296} & 0.234 & \textbf{0.292} & 0.239 & 0.299 & 0.245 & 0.305 & 0.229 & 0.301 & \underline{0.223} & 0.296 & 0.224 & 0.303 & 0.249 & 0.309 & 0.414 & 0.492 \\
 & 336 & \textbf{0.272} & \underline{0.330} & 0.294 & 0.339 & 0.299 & 0.388 & 0.295 & 0.346 & 0.286 & 0.341 & \underline{0.274} & \textbf{0.329} & 0.281 & 0.342 & 0.321 & 0.351 & 0.597 & 0.542 \\
 & 720 & \textbf{0.342} & \textbf{0.377} & 0.390 & 0.388 & 0.395 & 0.395 & 0.401 & 0.413 & 0.378 & 0.401 & \underline{0.362} & \underline{0.385} & 0.397 & 0.421 & 0.408 & 0.403 & 1.730 & 1.042 \\
 & Avg & \textbf{0.250} & \textbf{0.315} & 0.272 & \underline{0.317} & 0.277 & 0.322 & 0.279 & 0.329 & 0.284 & 0.339 & \underline{0.255} & \textbf{0.315} & 0.267 & 0.333 & 0.291 & 0.333 & 0.757 & 0.610 \\
\midrule
\multirow{5}{*}{\rotatebox{90}{Weather}} & 96  & \textbf{0.146} & 0.197 & 0.150 & \textbf{0.188} & 0.162 & 0.206 & 0.158 & \underline{0.195} & 0.162 & 0.212 & \underline{0.149} & 0.198 & 0.176 & 0.237 & 0.172 & 0.220 & 0.195 & 0.271 \\
 & 192 & \textbf{0.191} & \underline{0.240} & 0.202 & \textbf{0.238} & 0.211 & 0.250 & 0.211 & 0.247 & 0.204 & 0.248 & \underline{0.194} & 0.241 & 0.220 & 0.282 & 0.219 & 0.261 & 0.209 & 0.277 \\
 & 336 & \textbf{0.243} & \textbf{0.282} & 0.260 & \textbf{0.282} & 0.271 & 0.294 & 0.274 & 0.300 & 0.254 & 0.286 & \underline{0.245} & \textbf{0.282} & 0.265 & 0.319 & 0.280 & 0.306 & 0.273 & 0.332 \\
 & 720 & \textbf{0.313} & \textbf{0.333} & 0.343 & 0.353 & 0.345 & 0.343 & 0.401 & 0.413 & 0.326 & 0.337 & \underline{0.314} & \underline{0.334} & 0.333 & 0.362 & 0.365 & 0.359 & 0.379 & 0.401 \\
 & Avg & \textbf{0.223} & \underline{0.263} & 0.239 & \textbf{0.261} & 0.247 & 0.273 & 0.261 & 0.288 & 0.237 & 0.270 & \underline{0.225} & 0.263 & 0.248 & 0.300 & 0.259 & 0.287 & 0.264 & 0.320 \\
\bottomrule
\end{tabular}

\label{Table2}
\end{table*}
\subsection{Setups}
To maintain consistency with the baseline models, the training, validation, and test sets were normalized using the mean and standard deviation derived from the training data. Prediction horizons $H = \{96, 192, 336, 720\}$ are used for long-term forecasting tasks. In addition, for short-term forecasting tasks, the prediction lengths were chosen as follows: 6 for yearly, 8 for quarterly, and 18 for monthly.
\subsection{Key Results}
The results demonstrate the superior performance of the Times2D model compared to baseline forecasting models.
\subsubsection{Short-term forecasting:}
Table \ref{Table1} presents the short-term forecasting performance of Times2D across the M4 datasets. For the yearly data, Times2D achieved the lowest SMAPE at 13.418, surpassing advanced models like N-HITS (13.422) and PatchTST (13.477). This demonstrates the capability of Times2D to effectively capture long-term seasonal patterns. In the quarterly data, Times2D outperforms competing models such as TimesNet (10.465) and FEDformer (10.792) with a SMAPE of 10.362, emphasizing its robustness in managing seasonal variations. In addition, Times2D outperforms PatchTST (12.959) and N-HITS (13.059) in the monthly forecasting task, demonstrating its capability to forecast at higher temporal frequencies. 

\subsubsection{Long-term forecasting:} The long-term forecasting results are provided in Table \ref{Table2}. Times2D demonstrates superior performance in comparison to state-of-the-art models across multiple datasets and forecasting horizons $H = \{96, 192, 336, 720\}$. Specifically, Times2D achieves an average reduction of 5\% in MSE and 4\% in MAE across the datasets compared to advanced models like PatchTST and N-HITS. Times2D consistently provides lower error rates, particularly for longer forecasting horizons, with up to 8\% improvement in MSE and 7\% in MAE for the ETTm2 dataset at $H = 720$ compared to SageFormer, the best-second model. This highlights the strength of Times2D in capturing complex temporal dependencies over extended periods. The robustness of Times2D is further demonstrated by its superior performance over transformer-based models such as FEDformer and DLinear, achieving an average 6\% reduction in MSE across the Weather dataset. This highlights the effectiveness of Times2D in handling diverse data types. Additionally, comparisons with recent models like GPT4TS and CARD reveal that Times2D exceeds their performance, achieving a 7\% reduction in MAE on the ETTh1 dataset.
\begin{figure*}[t]
\centering
\raggedleft
\includegraphics[width=0.985\textwidth, keepaspectratio]{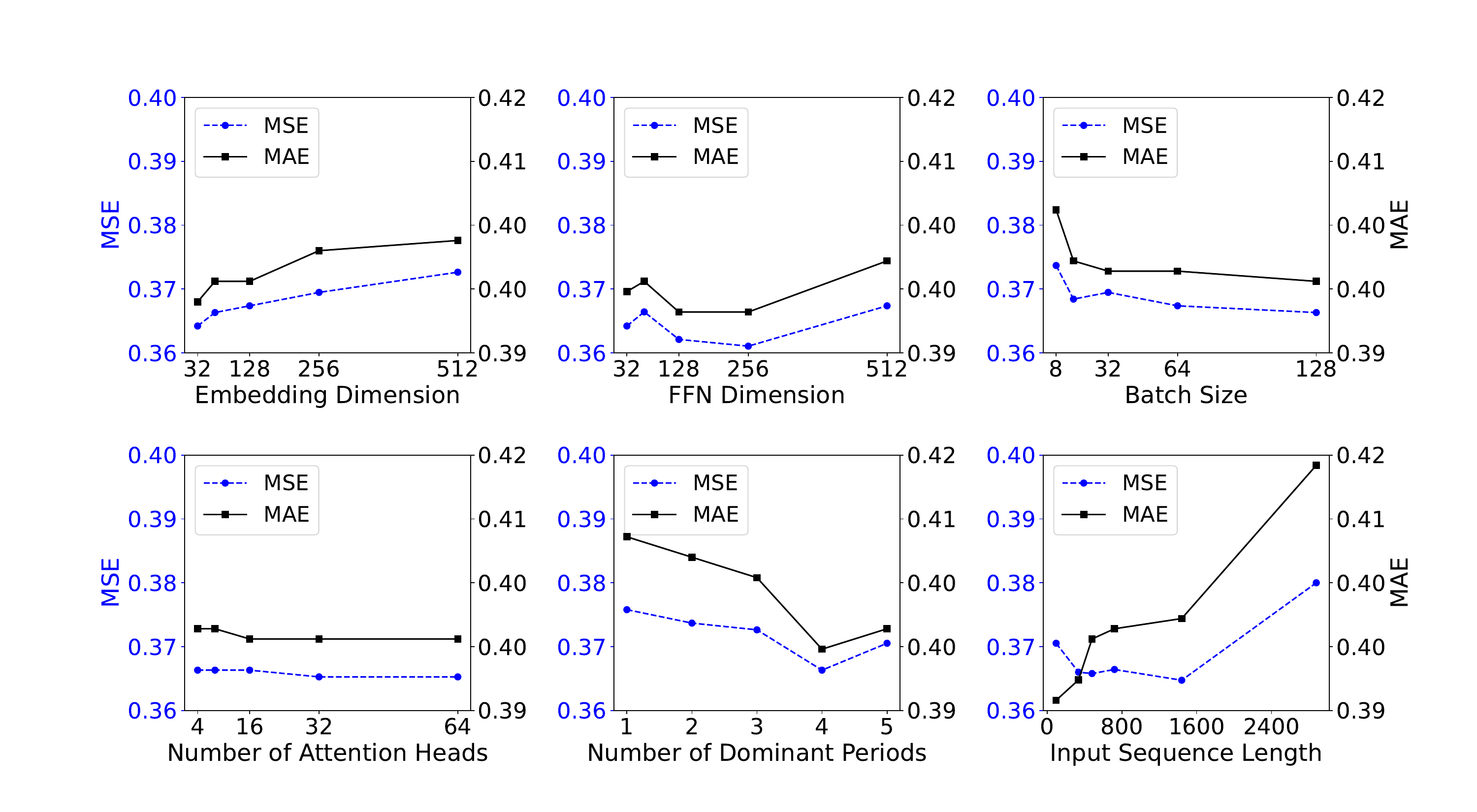} 
\caption{Sensitivity analysis of key hyperparameters for Times2D on the ETTh1 dataset with a prediction horizon of $H$=96.}
\label{Figure3}
\end{figure*}

\subsection{Computational Efficiency Analysis}
In this study, all simulations were conducted using PyTorch on a single NVIDIA L40S 46 GB GPU. The computational efficiency of the Times2D model was evaluated and compared with several state-of-the-art models, including TimesNet, PatchTST, Autoformer, Crossformer, and FEDFormer. To maintain consistency with the baseline models, we used average time per iteration and average RAM memory (MB) usage as key metrics across different prediction lengths $H= \{384,768,1536\}$. The results of this evaluation are summarized in Table \ref{Table3}. Times2D demonstrates a strong balance between computational speed and resource efficiency. The model maintains a consistently low average time per step across all prediction lengths, indicating its ability to scale effectively without significant performance degradation. In terms of memory usage, Times2D shows minimal variation, with RAM consumption ranging from 4329 MB to 4340 MB as prediction lengths increase. This low memory footprint highlights the suitability of Times2D for resource-constrained environments. In contrast, other models exhibited more pronounced increases in both computation time and memory usage with longer prediction lengths, underscoring the efficiency of Times2D in handling large-scale time series data. 
\begin{table}[t]
\centering
\fontsize{9}{10}\selectfont 
\caption{\setlength{\baselineskip}{10pt}\setlength{\linewidth}{\dimexpr\columnwidth-2cm\relax}
\small RAM memory usage (MB) and running time (S/iter) for prediction horizons $H = \{384, 768, 1536\}$.}
\setlength{\tabcolsep}{1.5mm} 
\renewcommand{\arraystretch}{1.2} 
\begin{tabular}{l|ccc|ccc}
\hline
Efficiency & \multicolumn{3}{c|}{RAM Memory (MB)} & \multicolumn{3}{c}{Running Time (S/iter)} \\ 
\hline
H & 384 & 768 & 1536 & 384 & 768 & 1536 \\ 
\hline
Times2D & 4329 & 4340 & 4324 & 0.045 & 0.050 & 0.052 \\ 
PatchTST & 6565 & 6805 & 8586 & 0.014 & 0.016 & 0.024 \\ 
TimesNet & 3908 & 3911 & 3903 & 0.018 & 0.025 & 0.079 \\ 
FEDformer & 18758 & 22341 & 28596 & 0.105 & 0.106 & 0.132 \\ 
Crossformer & 5674 & 5985 & 7272 & 0.022 & 0.023 & 0.028 \\ 
Autoformer & 6566 & 9914 & 16655 & 0.052 & 0.053 & 0.957 \\ 
\hline
\end{tabular}

\label{Table3}
\end{table}
\subsection{Sensitivity Analysis of Hyperparameters}
In this section, we examine key hyperparameters of the model, including embedding size, feedforward network size, batch size, attention heads, number of dominant periods, and input forecasting length. To do that, we conducted this analysis on ETTh1 data with a fixed forecasting horizon of 96, where only one hyperparameter changes while others remain fixed in each scenario. As shown in Figure~\ref{Figure3}, variations in the number of attention heads slightly affect model performance.  The model performs better with input lengths up to 1440. Beyond this, both MSE and MAE increase significantly, indicating that very long input sequences are ineffective. Lower embedding dimensions (32-64) and a feed-forward network size in the range of 128-256 provide better performance, while larger sizes gradually increase both MSE and MAE, raising the risk of overfitting. A batch size of 128 balances computational efficiency and predictive accuracy, with smaller batches resulting in higher errors and larger batches offering no significant improvement. 
\section{Conclusions}
In this paper, we introduced Times2D, an innovative and efficient framework for general time series forecasting that transforms 1D time series into 2D representations. This approach effectively captures complex temporal patterns, including multi-periodicity, sharp fluctuations, and turning points, which are often challenging for traditional 1D models. The proposed Periodic Decomposition Block (PDB) addresses multi-periodicity by converting the time series into a 2D tensor in the frequency domain, while the First and Second Derivative Heatmaps (FSDH) emphasize sharp changes and turning points. Extensive experiments on real-world datasets demonstrate that Times2D achieves superior performance and computational efficiency in both short-term and long-term forecasting, outperforming other state-of-the-art models. In the future, Times2D could be extended to address the detection of rare events, focusing on scenarios where anomalous or infrequent patterns affect forecasting accuracy significantly. Additionally, The model strengths and limitations can be examined in greater detail, rather than solely relying on aggregate metrics like MSE and MAE.
\section{Acknowledgments}
The authors declare financial support was received for the research and authorship of this article. This research was conducted with funding from NSF through award Number 2225341. 

\bibliographystyle{IEEEtran}
\bibliography{ref}

\end{document}